\documentclass[lettersize,journal]{IEEEtran}
\usepackage{amsmath,amsfonts}
\usepackage{algorithmic}
\usepackage{algorithm}
\usepackage{array}
\usepackage[caption=false,font=normalsize,labelfont=sf,textfont=sf]{subfig}
\usepackage{textcomp}
\usepackage{stfloats}
\usepackage{url}
\usepackage{verbatim}
\usepackage{graphicx}
\usepackage{cite}
\usepackage{booktabs}  
\usepackage{multirow}  
\usepackage{tabularx}
\usepackage{array}     
\usepackage{float}
\usepackage{soulpos}        
\usepackage{xcolor}         
\usepackage{enumitem}

\usepackage{hyperref}
\hypersetup{
	colorlinks=true,
	linkcolor=cyan,
	filecolor=blue,      
	urlcolor=black,
	citecolor=green,
}

\begin{document}

\title{DSFC-Net: A Dual-Encoder Spatial and Frequency Co-Awareness Network for Rural Road Extraction}

\author{Zhengbo Zhang, Yihe Tian, Wanke Xia, Lin Chen, Yue Sun, Kun Ding, Ying Wang, Bing Xu, \\
and Shiming Xiang 
\thanks{Zhengbo Zhang, Lin Chen, and Shiming Xiang are with the School of Artificial Intelligence, University of Chinese Academy of Sciences, Beijing 100049, China, and also with the State Key Laboratory of Multimodal Artificial Intelligence Systems, Institute of Automation, Chinese Academy of Sciences, Beijing 100190, China (email: zhangzhengbo2025@ia.ac.cn; chenlin2024@ia.ac.cn; smxiang@nlpr.ia.ac.cn).}
\thanks{Yihe Tian and Bing Xu are with the Department of Earth System Science, Tsinghua University, Beijing 100084, China (email: tianyh25@mails.tsinghua.edu.cn; bingxu@tsinghua.edu.cn)}
\thanks{Wanke Xia is with the Shenzhen International Graduate School, Tsinghua University, Shenzhen 518055, China (email: xwk25@mails.tsinghua.edu.cn).}
\thanks{Yue Sun is with the College of Information and Electrical Engineering, China Agricultural University, Beijing 100083, China (email: ndsy@cau.edu.cn).}
\thanks{Kun Ding and Ying Wang are with the State Key Laboratory of Multimodal Artificial Intelligence Systems, Institute of Automation, Chinese Academy of Sciences, Beijing 100190, China (email: kun.ding@ia.ac.cn; ywang@nlpr.ia.ac.cn).}
\thanks{\textit{Corresponding authors: Kun Ding and Bing Xu}}
\thanks{\textit{These authors contribute equally: Zhengbo Zhang and Yihe Tian}}
}


\maketitle

\begin{abstract}
Accurate extraction of rural roads from high-resolution remote sensing imagery is essential for infrastructure planning and sustainable development. However, this task presents unique challenges in rural settings due to several factors. These include high intra-class variability and low inter-class separability from diverse surface materials, frequent vegetation occlusions that disrupt spatial continuity, and narrow road widths that exacerbate detection difficulties. 
Existing methods, primarily optimized for structured urban environments, often underperform in these scenarios as they overlook such distinctive characteristics. To address these challenges, we propose DSFC-Net, a dual-encoder framework that synergistically fuses spatial and frequency-domain information. Specifically, a CNN branch is employed to capture fine-grained local road boundaries and short-range continuity, while a novel Spatial-Frequency Hybrid Transformer (SFT) is introduced to robustly model global topological dependencies against vegetation occlusions. Distinct from standard attention mechanisms that suffer from frequency bias, the SFT incorporates a Cross-Frequency Interaction Attention (CFIA) module that explicitly decouples high- and low-frequency information via a Laplacian Pyramid strategy. This design enables the dynamic interaction between spatial details and frequency-aware global contexts, effectively preserving the connectivity of narrow roads. Furthermore, a Channel Feature Fusion Module (CFFM) is proposed to bridge the two branches by adaptively recalibrating channel-wise feature responses, seamlessly integrating local textures with global semantics for accurate segmentation.
Comprehensive experiments on the WHU-RuR+, DeepGlobe, and Massachusetts datasets validate the superiority of DSFC-Net over state-of-the-art approaches. Notably, our method attains an F1-score of 69.93\% and an IoU of 53.77\% on the challenging WHU-RuR+ dataset, highlighting its robustness in rural scenes with narrow, sparse, and occluded roads. The code is available at \url{https://github.com/ZhengboZhang/DSFC-Net}.
\end{abstract}


\section{Introduction}
\label{sec:intro}
Rural roads constitute the backbone of agricultural infrastructure, bridging the gap between remote production areas and external markets. Facilitating this connectivity is a fundamental requirement for integrating rural populations into the modern economy\cite{weiss2018global}. The strategic significance of these networks is underscored by the fact that agrifood systems sustain the livelihoods of nearly half the global population, a demographic predominantly residing in rural regions\cite{rockstrom2020planet, verschuur2025heterogeneities}. Beyond serving as mere physical conduits, these routes act as vital arteries for the exchange of information, technology, and capital. Consequently, the continuous monitoring of rural road networks is indispensable for the Sustainable Development Goals (SDGs)\cite{burke2021using}, serving as a key indicator for eradicating poverty (SDG 1), ending hunger (SDG 2), and fostering resilient infrastructure (SDG 9)\cite{koks2019global}.

Accurate road extraction forms the foundation for effective network monitoring. Traditional approaches primarily depended on handcrafted features and geometric modeling. Techniques such as mathematical morphology\cite{soille2002advances, valero2010advanced} and knowledge-based reasoning\cite{hinz2003automatic, liu2017rural}     were widely employed to identify road-like structures based on linearity and width consistency. However, these methods often required manual parameter tuning and lacked the generalization capability needed for complex environments\cite{mena2003state}. In contrast, over the past decade, deep learning has become the dominant paradigm for this task. Early methods mainly relied on Convolutional Neural Networks (CNNs) to extract image features\cite{dlinknet,zhang2018road,9226530}. However, CNNs often struggle to capture long-range dependencies due to their local receptive fields. To address this, recent research has shifted toward Transformers. These architectures utilize self-attention mechanisms to effectively model global context\cite{liu2021swin, cao2022swin,10539008}. Most recently, the focus has turned to tackling complex geometric challenges, such as occlusion and topological connectivity, using specialized modules\cite{he2020sat2graph, denet, segroadv2}.


\begin{figure}[t] 
  \centering
  \includegraphics[width=\linewidth]{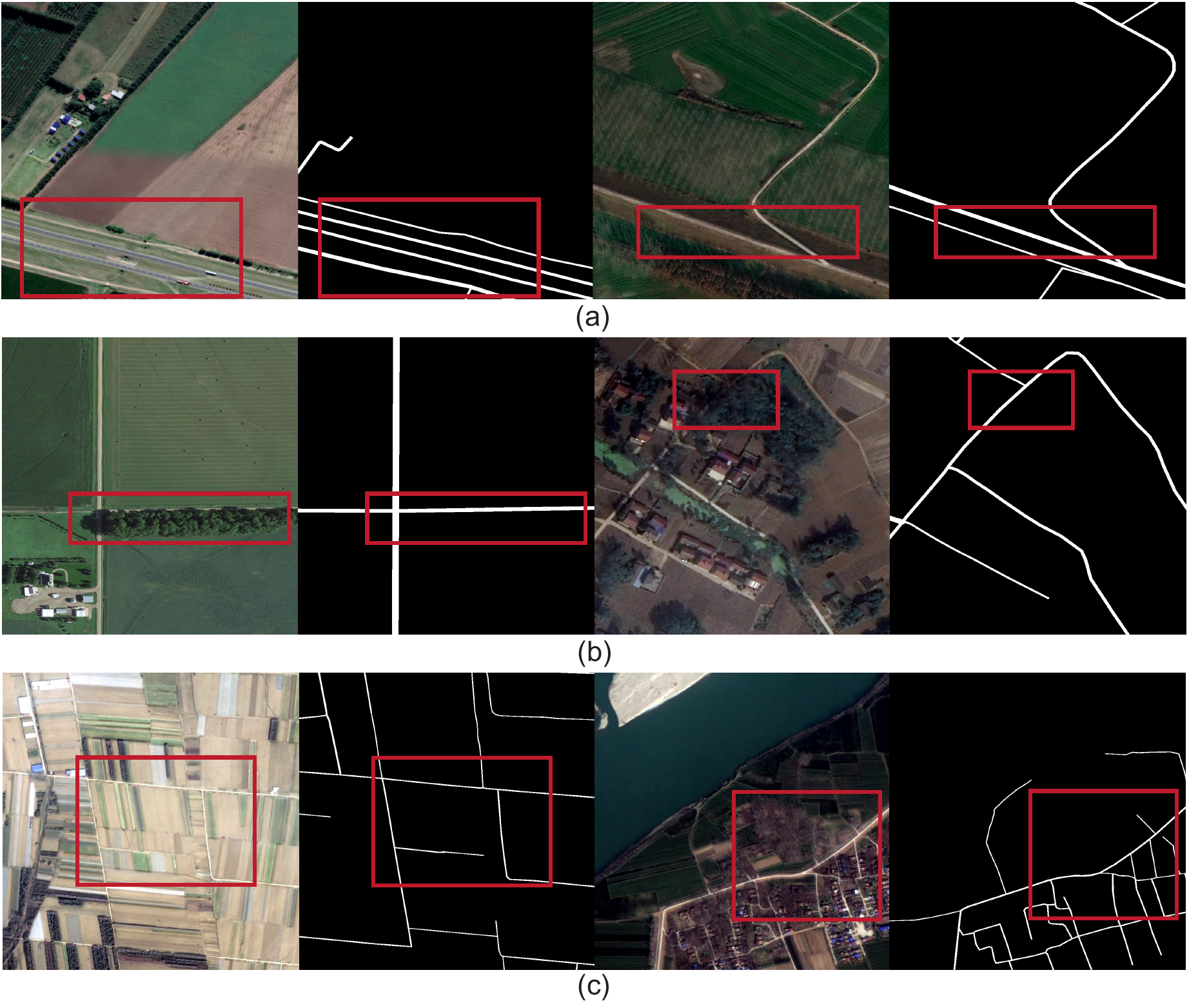}
  \caption{Illustrations of the three primary challenges in extracting rural roads from remote sensing imagery. (a) High intra-class variability and low inter-class separability, where diverse road surfaces blend with surrounding bare soil or fields. (b) Occlusion by vegetation, disrupting road continuity due to trees and crops. (c) Narrow roads, resulting in detection difficulties and foreground-background imbalance.}
  \label{fig1}
\end{figure}    


However, most existing methods are designed for urban scenes and often fail in rural settings. 
This performance gap stems from the significant differences between rural and urban road networks. Specifically, rural roads present three unique challenges, as illustrated in Fig. \ref{fig1}. First, they exhibit high intra-class variability and low inter-class separability (Fig. ~\ref{fig1}(a)): diverse surface materials lead to varying spectral features within the road class, while unpaved roads can be easily confused with background elements such as bare soil or field ridges. Second, rural roads are frequently occluded by vegetation, including trees and crops, which disrupts their spatial continuity and complicates accurate extraction (Fig. ~\ref{fig1}(b)). Finally, rural roads are typically narrow, leading to greater detection difficulty and an imbalance between foreground and background elements that hinders detection in remote sensing imagery (Fig. ~\ref{fig1}(c)) \cite{rurp}.

To tackle these challenges, we propose DSFC-Net, a dual-encoder framework that synergistically leverages the complementary strengths of spatial and frequency-domain processing. Specifically, the CNN branch effectively extracts local road features to mitigate classification inaccuracies caused by diverse surface materials, while the Spatial-Frequency Hybrid Transformer (SFT) branch, equipped with Cross-Frequency Interaction Attention (CFIA), explicitly decouples frequency components to model global topological dependencies and isolate road boundaries from complex background noise. To bridge these two branches, a Channel Feature Fusion Module (CFFM) is introduced to adaptively recalibrate channel-wise feature responses, seamlessly aligning local details with global semantics. Comprehensive experiments on the WHU-RuR+, DeepGlobe, and Massachusetts datasets validate that DSFC-Net achieves state-of-the-art performance, demonstrating superior robustness in parsing narrow, sparse, and occluded rural roads.

The main contributions of this paper are summarized as follows:
\begin{enumerate}
  \item We propose DSFC-Net, a novel dual-encoder based network specifically designed for rural road extraction from remote sensing images, which models the spatial and frequency domain information to address the unique challenges in rural road extraction.
  \item We propose a Spatial-Frequency Hybrid Transformer (SFT) with a Cross-Frequency Interaction Attention (CFIA) module. By decomposing high and low frequency features, it captures long-range dependencies and addresses challenges of narrow roads and spatial discontinuities. Additionally, a Channel Feature Fusion Module (CFFM) bridges the CNN and Transformer branches through adaptive channel weight recalibration.
  \item Extensive experiments on three public datasets, characterized by varying rural complexities and sparse road densities, demonstrate that DSFC-Net achieves state-of-the-art performance. Notably, our method achieves an F1-score of 69.93\% and an IoU of 53.77\% on the WHU-RuR+ dataset, highlighting its superior accuracy and robustness in challenging rural scenarios.

\end{enumerate}

\section{Related Work}

\subsection{Road Extraction Methods}
Deep learning has fundamentally transformed the paradigm of road extraction, evolving from basic Convolutional Neural Networks (CNNs) to sophisticated architectures integrating attention mechanisms and topology-aware modules. Early approaches primarily focused on optimizing encoder-decoder structures to enhance feature representation. For instance, ResU-Net\cite{zhang2018road} integrates residual blocks into the U-Net backbone to facilitate deep feature learning, while the seminal D-LinkNet\cite{dlinknet} employs dilated convolutions to significantly expand the receptive field without reducing spatial resolution. However, these CNN-based methods are inherently limited by their local receptive fields. In rural scenarios characterized by high intra-class variability, they often struggle to distinguish subtle road features from the complex background, leading to fragmented extraction results.

To overcome the inherent limitation of CNNs in capturing long-range dependencies, subsequent research introduced attention mechanisms and global perception modules. GAMSNet\cite{lu2021gamsnet} utilizes multi-scale residual learning to capture intricate spatial and channel-wise dependencies. Marking a shift toward transformer-based architectures, Swin-Unet\cite{cao2022swin} leverages shifted window self-attention to model global context effectively. Most recently, researchers have focused on efficient context modeling. CGCNet\cite{cgcnet} introduces a compact global context-aware block to capture long-range dependencies with reduced computational complexity. CFRNet\cite{cfrnet} boosts segmentation accuracy by fusing rich features across multiple scales via a cascade architecture, while FRCFNet\cite{frcfnet} proposes a grouping multidimensional feature reassembly module, utilizing strip convolutions to fuse context information across various directions. Despite effective global modeling, standard attention mechanisms exhibit frequency bias by prioritizing low-frequency signals. This results in over-smoothed boundaries and the failure to preserve fine-grained road details in complex rural scenes.

The research focus has further transitioned toward refining boundaries and preserving topological connectivity, particularly to address road fragmentation and occlusion. BT-RoadNet\cite{zhou2020bt} adopts a coarse-to-fine strategy to improve boundary quality, while Sat2Graph\cite{he2020sat2graph} directly extracts road graphs to ensure geometric correctness. To tackle complex occlusions, OARENet\cite{oarenet} designs a specialized occlusion-aware decoder that employs strip convolutions to recover road connectivity in dense urban environments. Furthermore, DENet\cite{denet} emphasizes the importance of directional features by integrating a road edge detector with directional spatial attention to trace narrow and elongated roads. Finally, SegRoadv2\cite{segroadv2} innovatively combines deformable self-attention with groupable deformable convolutions to adaptively trace road trajectories. However, most topology-aware approaches rely on geometric priors tailored for structured urban environments. Consequently, they often fail to generalize to the winding and irregular geometry of rural roads and struggle to recover continuity under unstructured canopy occlusions.

Despite their dominance in urban datasets, these methods suffer from a precipitous decline in performance when applied to rural imagery. They struggle to generalize to the complex rural reality, where the extreme textural heterogeneity and sparse road connectivity render existing urban-focused priors ineffective. To tackle these challenges, our proposed DSFC-Net adopts a dual-encoder strategy that synergistically fuses CNN and Transformer branches. By explicitly decoupling high-frequency boundary details from low-frequency topological dependencies via a Spatial-Frequency hybrid Transformer, our method effectively mitigates background interference and preserves the continuity of narrow, winding rural roads.

\subsection{Rural Road Extraction}
Compared to structured urban road networks, rural roads present unique and severe challenges: they are typically much narrower, exhibit high intra-class variance due to diverse surface materials, and suffer from heavy occlusion by vegetation. High-quality datasets serve as the cornerstone of data-driven methods, yet classic datasets have predominantly focused on urban areas, leading to models that generalize poorly to complex rural environments. To bridge this data gap, recent efforts have been dedicated to constructing rural-specific datasets. Notably, WHU-RuR+\cite{rurp} provides a large-scale dataset covering 6,866 km² across eight countries, explicitly designed to represent the heterogeneous backgrounds of global rural areas. Similarly, the WHU-CR dataset\cite{wang2026identifying} was constructed to capture the diverse geographic and environmental contexts specific to China's rural regions.

Despite the emergence of these datasets, specialized algorithms dedicated to rural road extraction remain remarkably scarce. Among the few pioneering works, R2-Net\cite{wang2024reverse} introduces a reverse refinement mechanism designed to extract narrow roads by enhancing boundary separability. To further mitigate feature loss during the deep encoding of narrow targets, NANet\cite{wang2026identifying} employs a coarse-to-fine framework integrating a foreground-background collaborative mechanism. 

However, research in rural scenes is still in its infancy. The limited number of existing studies often struggles to simultaneously handle the extreme scale variations and high-frequency noise inherent in rural remote sensing images. Consequently, there is an urgent need for more robust, dedicated architectures capable of navigating these complex scenarios, a gap that this study aims to fill.

\begin{figure*}[t] 
  \centering
  \includegraphics[width=\linewidth]{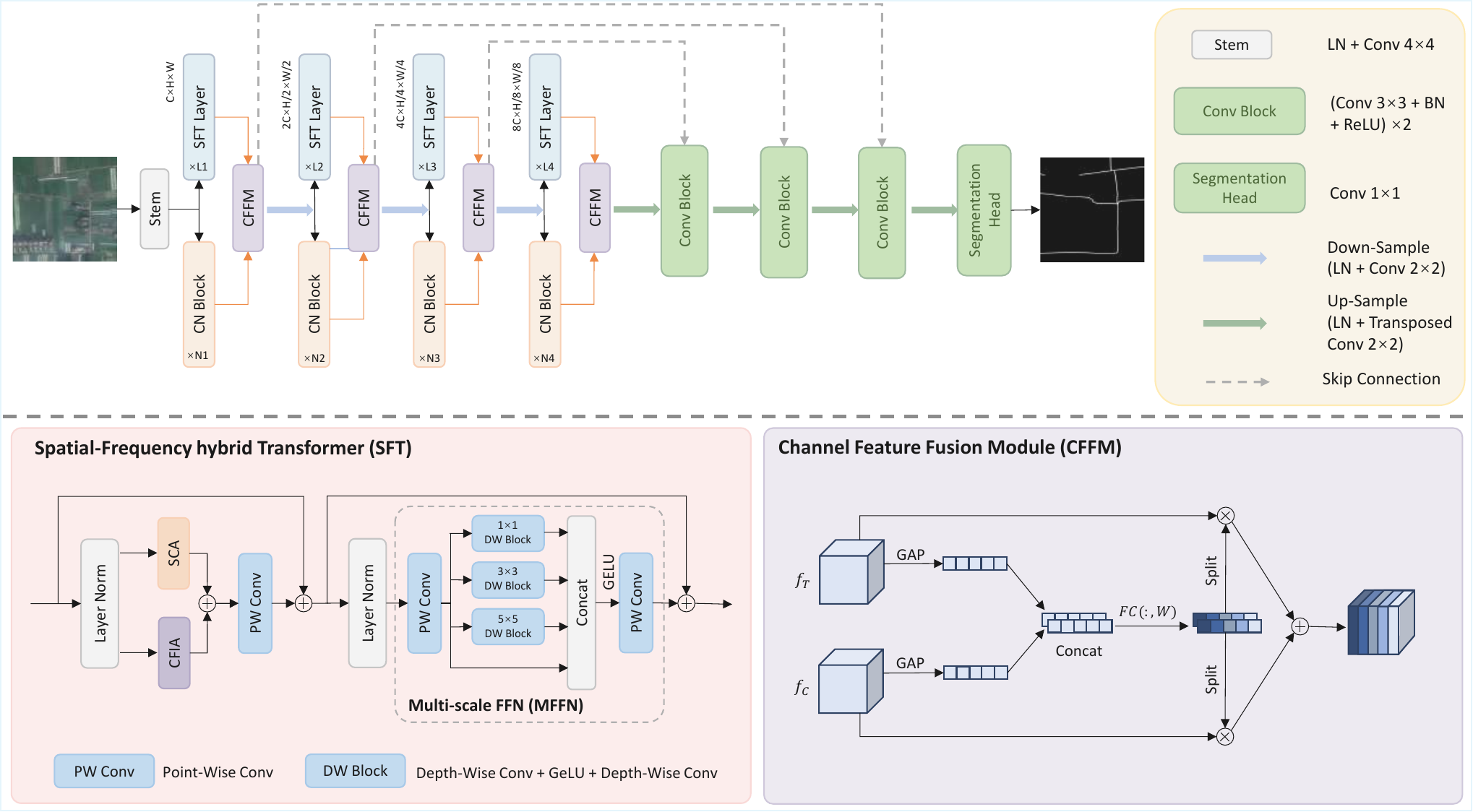}
  \caption{Overall architecture of the proposed DSFC-Net. The structure of DSFC-Net consists of two parallel encoders, a CNN branch using ConvNeXt-v2 \cite{convnextv2} and a Transformer branch using our proposed SFT. And the CFFM is used to fuse the output feature maps of the two encoders.}
  \label{fig2}
\end{figure*}

\section{Method}

\subsection{Overall Architecture}

The overall architecture of DSFC-Net is shown in Fig. \ref{fig2}, where we use both the CNN branch and the Transformer branch to effectively extract spatial and frequency features respectively.
Given an input image $x \in \mathbb{R}^{H \times W \times 3}$, DSFC-Net first appiles a stem block to extract low-level feature maps $F_0 \in \mathbb{R}^{H \times W \times C}$, where $H \times W$ denotes the spatial dimensions and $C$ is the number of channels. Next, the feature maps $F_0$ are passed through a 4-stage encoder structure sequentially. Between each stage, CNN and Transformer branches share a downsampling module (Layer Normalization and a $2\times2$, stride 2 convolution layer). Inside each stage, the input features are passed through $N_i$ CNN Blocks and $L_i$ SFT layers to extract feature maps $F^i_C$ and $F^i_T$ separately. 
These two features are fused by the Channel Feature Fusion Module (CFFM) to obtain low-level feature maps $F^i_l$, where $i \in \{1,2,3,4\}$. Then, transposed convolutions are applied for upsampling in the decoder, and the low-level feature maps $F^i_l$ are concatenated with the high-level feature maps $F^i_h$ via skip connection. Finally, the feature maps are magnified 4 times to their original size through a transposed convolution, and the final segmentation mask
is obtained through a segmentation head.

The DSFC-Net has two key components: the Spatial-Frequency Hybrid Transformer (SFT) is used as a Transformer branch to perceive global features, and the Channel Feature Fusion Module (CFFM) is used to fuse the feature maps of the two branches. We will introduce these two components in the following subsections.

\subsection{Spatial-Frequency Hybrid Transformer (SFT)}

While CNN branch excels at extracting local detail features, rural road extraction tasks often face challenges such as the continuous and slender structure of roads, which require strong global context modeling. To address this, we propose a Spatial-Frequency Hybrid Transformer (SFT) that captures the complex spatial structure and frequency features in rural road scenes. This module has three components: Spatial Context Aggregator (SCA), Cross-Frequency Interaction Attention (CFIA) mechanism and a Multi-scale Feed-Forward Network (MFFN). These components integrate local spatial perception capabilities with global frequency representation characteristics, enhancing the ability to express road edges, texture changes, and long-distance contextual information.

As shown in Fig. \ref{fig2}, the SFT employs a parallel architecture to process spatial and frequency information simultaneously. Given input features $F^i_l \in \mathbb{R}^{H \times W \times C}$, they first undergo Layer Normalization (LN), and then are fed into Spatial Context Aggregator (SCA) and Cross-Frequency Interaction Attention (CFIA) branches respectively. The outputs of these two branches are aggregated, processed by a point-wise convolution, and added to the input via a residual connection. This process can be denotes as:

\begin{equation}
    \tilde{F^i_l} = F^i_l + \text{PW}_{conv}(\text{SCA}(\text{LN}(F^i_l))+\text{CFIA}(\text{LN}(F^i_l)))
\end{equation}

Then, the aggregated features are fed into a Multi-scale Feed-Forward Network (MFFN) to further extract multi-scale semantic information. The output of the MFFN is then processed through residual connections to obtain the final output of the SFT.

\subsubsection{Spatial Context Aggregator (SCA)}
Road objects exhibit extreme scale variations, but a fixed-size receptive field is insufficient to capture these diverse geometric structures simultaneously, making it difficult for long roads to maintain connectivity. To bridge this gap, we introduce the Spatial Context Aggregator (SCA) module. This module employs parallel depth-wise dilated convolutions with different dilation rates to capture and aggregate multi-scale contextual cues. 

The first branch utilizes a standard $3 \times 3$ depth-wise convolution with a dilation rate $d=1$. This branch focuses on extracting fine-grained local details and preserving the continuity of road surfaces in a compact neighborhood. The second branch employs a $3 \times 3$ depth-wise convolution with a dilation rate $d=2$. By expanding the receptive field without reducing spatial resolution, this branch captures broader contextual information, which is essential for recognizing road topology and connecting disjointed road segments. Then, the concatenated features are processed by a fusion convolution block designed to facilitate cross-channel interaction and dimensional reduction.



\subsubsection{Cross-Frequency Interaction Attention (CFIA)}

\begin{figure}[htbp] 
  \centering
  \includegraphics[width=\linewidth]{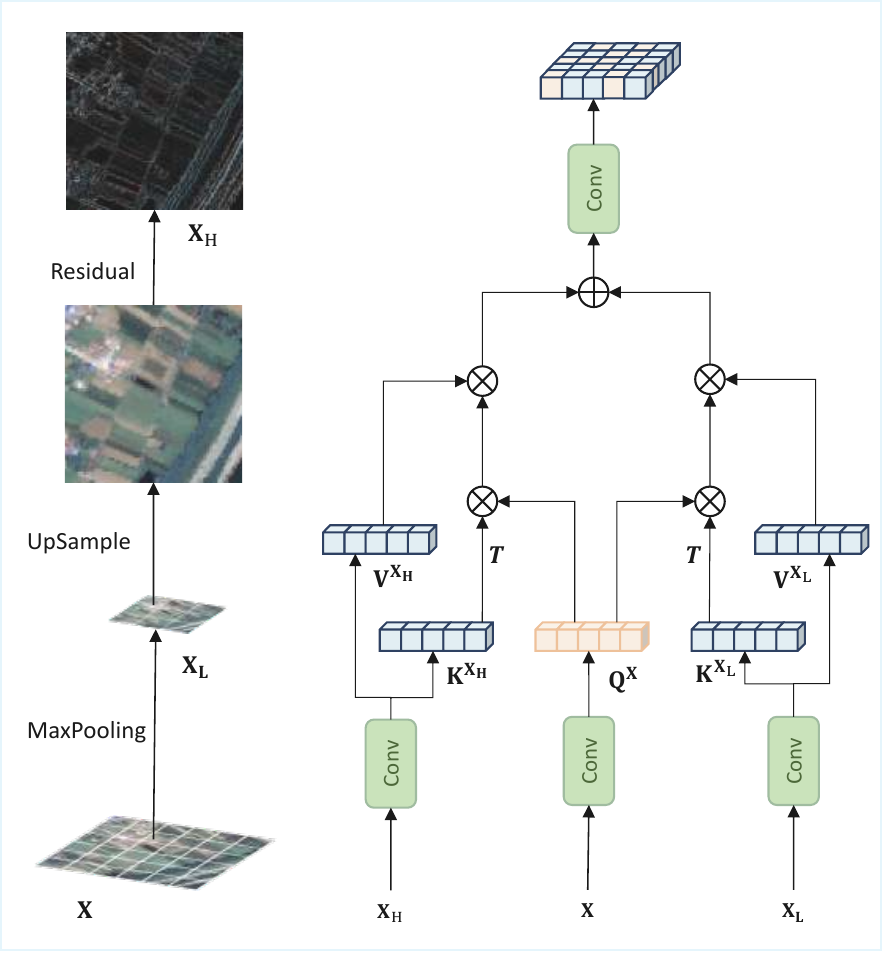}
  \caption{The structure of proposed Cross-Frequency Interaction Attention (CFIA). \textit{Left}: The example of Laplacian Pyramid decomposition strategy. \textit{Right}: The detail structure of CFIA, where $\mathbf{X}_{L}$ and $\mathbf{X}_{H}$ are obtained from Laplacian Pyramid decomposition strategy.}
  \label{fig3}
\end{figure}

In road segmentation, the semantic information exists in orthogonal frequency domains: the topological structure and connected road regions exhibit smooth intensity variations and are predominantly characterized by low-frequency components, whereas road boundaries, lane markings, and pavement cracks involve sharp intensity transitions and thus correspond to high-frequency components. Standard self-attention mechanisms in Transformer operate on mixed signals, often leading to a \textit{frequency bias} where high-frequency details are overwhelmed by dominant low-frequency patterns. 
This bias results in blurred segmentation boundaries and disconnected road topologies. To address this limitation, we propose the Cross-Frequency Interaction Attention (CFIA) module. As shown in Fig. \ref{fig3}, our CFIA module explicitly disentangles the input feature $\mathbf{X} \in \mathbb{R}^{H \times W \times C}$ into two distinct frequency components. 


Instead of relying on computationally expensive Fourier transforms, we employ a more efficient, parameter-free Laplacian Pyramid decomposition strategy, enabling targeted processing of global contexts. The low-frequency features $\mathbf{X}_{L}$ are obtained via a max-pooling operation with stride $s$, acting as a structural low-pass filter, preserving the main road layout while effectively suppressing noise and reducing spatial resolution for efficient global modeling. The high-frequency component $\mathbf{X}_{H}$, which encapsulates edge and texture information, is derived from the residual difference between the upsampled low-frequency features and the original input. This decomposition is formulated as:

\begin{equation} 
    \mathbf{X}_{L} = MaxPool_{s}(\mathbf{X}),
\end{equation}

\begin{equation}
    \mathbf{X}_{H} = UpSample(\mathbf{X}_{L}) - \mathbf{X}
\end{equation}

The core innovation of CFIA lies in its interaction strategy. Rather than computing self-attention within each isolated frequency domain, we employ the original feature $\mathbf{X}$ after convolution projection as a semantic query anchor. This design ensures that the retrieved frequency-specific information remains aligned with the original semantic context. We generate the Query vector $\mathbf{Q}^{\mathbf{X}}$ from $\mathbf{X}$, while obtaining Key and Value vectors from the decomposed components: $\mathbf{K}^{\mathbf{X}_H}, \mathbf{V}^{\mathbf{X}_H}$ from the high-frequency branch, and $\mathbf{K}^{\mathbf{X}_L}, \mathbf{V}^{\mathbf{X}_L}$ from the low-frequency branch. This dual-branch structure allows for complementary feature refinement. The high-frequency Branch focuses on local detail restoration. By querying the high-frequency key-value pairs, the model attends to sharp transitions, thereby refining road boundaries. The low-frequency Branch exploits the reduced resolution of $\mathbf{X}_{L}$ to capture long-range dependencies and global topology. Then, we calculate the multi-head attention for both branches:

\begin{equation}
\text{Attn}^{high}_{i} = Softmax\left( \frac{\mathbf{Q}^{\mathbf{X}}_i \cdot (\mathbf{K}^{\mathbf{X}_H}_i)^{\mathbf{\textit{T}}}}{\sqrt{d_k}} \right) \cdot \mathbf{V}^{\mathbf{X}_H}_i,
\end{equation}

\begin{equation}
\text{Attn}^{low}_{i} = Softmax\left( \frac{\mathbf{Q}^{\mathbf{X}}_i \cdot (\mathbf{K}^{\mathbf{X}_L}_i)^{\mathbf{\textit{T}}}}{\sqrt{d_k}} \right) \cdot \mathbf{V}^{\mathbf{X}_L}_i
\end{equation}

\noindent where $i$ denotes the index of heads, $d_k$ denotes the dimension of each head. Then, the outputs from all the heads are concatenated. Finally, the high-frequency features and low-frequency features are aggregated and projected back to the original dimension, yielding a comprehensive frequency-aware representation. This comprehensive frequency-aware representation effectively integrates the global topological consistency from the low-frequency features with the precise boundary details from the high-frequency features.

\subsubsection{Multi-scale Feed-Forward Network (MFFN)}

For road segmentation tasks, distinguishing between road surfaces and backgrounds relies heavily on local texture continuity and varying structural scales, so we propose the Multi-scale Feed-Forward Network (MFFN) to extract local relationships for high-dimension features from various receptive fields.

Given the input feature from the preceding module, we first reduce the channel dimension using a point-wise convolution, generating an intermediate feature map. To capture diverse spatial patterns, the intermediate feature map is then processed through parallel depth-wise convolution blocks with varying kernel sizes. The small kernel size branch preserves the fine-grained, pixel-level information, and large kernel size branch perceives larger structural patterns, such as wide lane markings. Additionally, to facilitate efficient gradient flow and feature reuse, we include an identity path that directly forwards the expanded features.

\subsection{Channel Feature Fusion Module (CFFM)}

In our dual-branch architecture, using direct fusion strategies such as simple element-wise summation or concatenation would lead to treating the features from these distinct domains equally, thus ignoring the dynamic variation in their informativeness. To address this, we propose the Channel Feature Fusion Module (CFFM), the structure is shown in Fig. \ref{fig2}. Inspired by the Squeeze-and-Excitation (SE) mechanism \cite{seblock}, CFFM is designed to explicitly model the inter-dependencies between the two branches and adaptively recalibrate channel-wise feature responses.

Given the input feature maps $F_C \in \mathbb{R}^{H \times W \times C}$ and $F_T \in \mathbb{R}^{H \times W \times C}$ from the two parallel branches. The fusion process begins with a joint squeeze operation to aggregate global spatial information. We apply Global Average Pooling (GAP) to both inputs independently and concatenate their channel descriptors to form a holistic context vector $\mathbf{z}_{joint} \in \mathbb{R}^{2C}$. This joint vector encapsulates the global distribution of the entire dual-branch system. To capture the non-linear interactions between channels across the two branches, $\mathbf{z}_{joint}$ is passed through a MLP bottleneck structure. Then, the attention weights $\mathbf{s} = \sigma( MLP (\mathbf{z}_{joint}) )$ are derived from the concatenated representation through a Sigmoid gate, effectively reflecting the importance score of cross-branch information. 

Finally, the generated attention vector $\mathbf{s}$ is split into two sub-vectors $\mathbf{s}_C \in \mathbb{R}^{C}$ and $\mathbf{s}_T \in \mathbb{R}^{C}$, corresponding to the two input branches respectively. These weights serve as soft gates to recalibrate the original input features. The final fused output $F_{l}$ is obtained by the weighted summation of the The final fused output $F_{l}$ is obtained by the weighted summation of the recalibrated features. 



\begin{table*}[htbp]
\normalsize
\centering
\caption{Quantitative comparison of the proposed DSFC-Net with the state-of-the-art road extraction methods on the WHU-RuR+ dataset, in which bold is the best and underline is the second-best.}
\label{tab:main1}
\begin{tabular*}{0.7\textwidth}{@{\extracolsep{\fill}}cccccc@{}}
\toprule
\textbf{Methods} &  \textbf{Source}  & \textbf{Precision (\%)} & \textbf{Recall (\%)} & \textbf{F1 (\%)} & \textbf{IoU (\%)} \\ 
\midrule
U-Net \cite{unet} & $\text{MICCAI}_{2015}$ & 76.18 & 58.87 &  66.42  &  49.72  \\
D-LinkNet \cite{dlinknet} & $\text{CVPR}_{2018}$ & 72.46 & 65.88 &  69.01  &  52.68  \\
SegFormer \cite{segformer} & $\text{NIPS}_{2021}$  &  75.91  &  60.54  &  67.36  &  50.78  \\
CFRNet \cite{cfrnet} & $\text{GRSL}_{2024}$ &   79.01  &  57.15  &  66.33  &  49.62  \\
FRCFNet \cite{frcfnet} & $\text{GRSL}_{2024}$  &  81.18  &  48.83  &  60.98 &  43.86   \\
OARENet \cite{oarenet} & $\text{TGRS}_{2024}$  & 77.09 & 59.18 & 66.96  & 50.33  \\
$\text{$\text{C}^2$Net}$ \cite{ccnet} & $\text{TGRS}_{2024}$  & 80.03 & 58.45 &  67.56  &  51.01  \\
R2-Net \cite{wang2024reverse} & $\text{arxiv}_{2024}$  &  65.49  &    73.79    &  \underline{69.40}  &   \underline{53.14}  \\
DENet \cite{denet} & $\text{TITS}_{2025}$  &  71.89  &     56.90  &  63.52  &  46.54 \\
SegRoadv2 \cite{segroadv2} & $\text{IJDE}_{2025}$  & 74.48 & 61.54 &  67.39  & 50.82 \\ 
CGCNet \cite{cgcnet} & $\text{TGRS}_{2025}$  & 77.19 & 60.84 &  68.04  &  51.57  \\
\midrule
\textbf{DSFC-Net(Ours)} &    &  75.77  &  64.93  &  \textbf{69.93}  &   \textbf{53.77}  \\ 
\bottomrule
\end{tabular*}
\end{table*}

\begin{figure*}[htbp] 
  \centering
  \includegraphics[width=\linewidth]{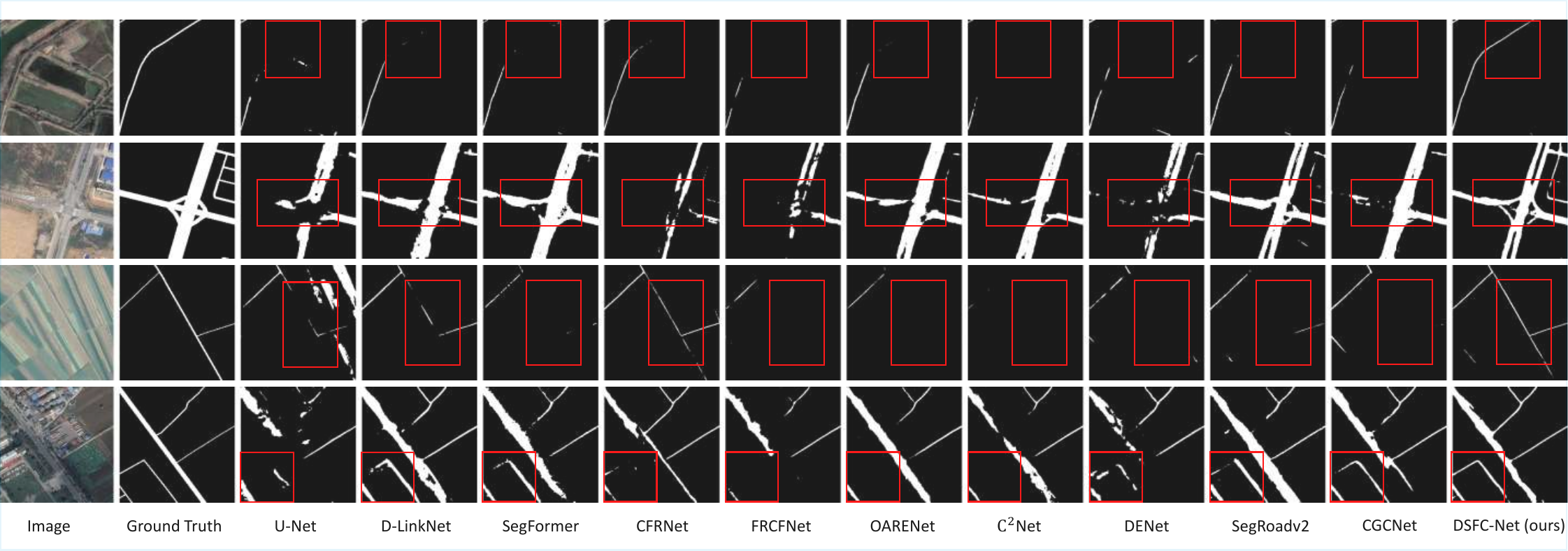}
  \caption{Visual comparison of the different methods on the WHU-RuR+ dataset.}
  \label{fig4}
\end{figure*}

\section{Experiment and Analysis}

\subsection{Datasets}
    In our experiments, we use three datasets to evaluate our method: WHU-RuR+\cite{rurp}, DeepGlobe\cite{deepglobe}, and  Massachusetts\cite{mass}. These datasets are ideal for simulating rural road scenarios. They differ significantly from standard urban datasets because they have a high percentage of rural backgrounds and a very low density of road pixels. Specifically, the rural area coverage is 98.4\% for WHU-RuR+, 74.31\% for DeepGlobe, and 49.65\% for Massachusetts. Meanwhile, the road pixel ratio in all three datasets is less than 5\%. This data distribution authentically reflects the challenge of rural areas characterized by the difficulty of identifying narrow and sparse roads within vast and complex backgrounds.

\subsubsection{WHU-RuR+} This dataset is a large-scale dataset tailored for rural road extraction, comprising 36,098 image pairs ($1024 \times 1024$) with a spatial resolution of 0.3–0.8m. Covering 6,866.35 $\text{km}^2$ across eight countries, it captures the unique complexity of rural scenes, including varying widths and unpaved surfaces. We follow the standard partition of 18,103 images for training and 17,995 for testing.

\subsubsection{DeepGlobe} Originating from the DeepGlobe 2018 Challenge, this dataset provides sub-meter (0.5 m) satellite imagery ($1024 \times 1024$) covering Thailand, Indonesia, and India. It features diverse road types and illumination conditions across a total area of 2,220 $\text{km}^2$. In this study, we utilized the official training set of 4,980 images for training. Due to the unavailability of ground truth labels for the official test set, we employed the entire official validation set, consisting of 1,236 images, as our testing set for evaluation.

\subsubsection{Massachusetts} This dataset contains 1,171 aerial images ($1500 \times 1500$) with a 1-meter resolution, covering over 2,600 $\text{km}^2$ of diverse urban and rural terrain. Following standard protocols, the dataset is strictly divided into 1,108 images for training and 63 images for testing.

\subsection{Implement Details}

In the DSFC-Net, we apply ConvNeXt-v2 \cite{convnextv2} blocks for the CNN branch and the proposed Spatial-Frequency Hybrid Transformer (SFT) layers for the Transformer branch. The experiments are conducted on 4 NVIDIA RTX 3090 GPUs with 24GB VRAM. For the WHU-RuR+ dataset, the original image is resized to 768$\times$768 pixels. For the DeepGlobe dataset, the original image is divided into four 512$\times$512 patches. For the Massachusetts dataset, the original image is divided into nine 500$\times$500 patches, and resized to 512$\times$512 pixels. During the training phase, we use some data augmentation methods to avoid overfitting, including horizontal flip, vertical flip and diagonal flip. All models are trained using AdamW optimizer, with a batch size of 4 over 100 epochs, and the learning rate is initially set at 1e-4. The loss function is the sum of binary cross-entropy (BCE) loss and dice loss.

\begin{table*}[htbp]
\normalsize
\centering
\caption{Quantitative comparison of the proposed DSFC-Net with the state-of-the-art road extraction methods on the DeepGlobe dataset, in which bold is the best and underline is the second-best.}
\label{tab:main2}
\begin{tabular*}{0.7\textwidth}{@{\extracolsep{\fill}}cccccc@{}}
\toprule
\textbf{Methods} &  \textbf{Source}  & \textbf{Precision (\%)} & \textbf{Recall (\%)} & \textbf{F1 (\%)} & \textbf{IoU (\%)} \\ 
\midrule
U-Net \cite{unet} & $\text{MICCAI}_{2015}$  &  84.59  &   73.81   &  78.83  &   65.06  \\
D-LinkNet \cite{dlinknet} & $\text{CVPR}_{2018}$  &  82.65  &   76.60   &  79.54  &   65.99  \\
SegFormer \cite{segformer} & $\text{NIPS}_{2021}$  &  82.10   &   76.91   &  79.42  &   65.86  \\
CFRNet \cite{cfrnet} & $\text{GRSL}_{2024}$  &   82.28  &    78.46    &  \underline{80.32}  &   \underline{67.12}  \\
FRCFNet \cite{frcfnet} & $\text{GRSL}_{2024}$  &   83.33  &   72.71   &  77.66  &   63.48  \\
OARENet \cite{oarenet} & $\text{TGRS}_{2024}$  &  83.59  &    76.65    &  79.97  &   66.63  \\
$\text{$\text{C}^2$Net}$ \cite{ccnet} & $\text{TGRS}_{2024}$  &  82.94  &    73.44    &  77.90  &  63.80  \\
DENet \cite{denet} & $\text{TITS}_{2025}$  &  73.92  &    84.25    &  78.75  &   64.95  \\
SegRoadv2 \cite{segroadv2} & $\text{IJDE}_{2025}$  &  83.55  &  76.55  &  79.90  &  66.52  \\ 
CGCNet \cite{cgcnet} & $\text{TGRS}_{2025}$  &  82.89  &  76.87  &  79.77  &  66.35 \\
\midrule
\textbf{DSFC-Net(Ours)} &  &  82.68  &  80.23  &  \textbf{81.44}  &  \textbf{68.68}  \\ 
\bottomrule
\end{tabular*}
\end{table*}

\begin{figure*}[htbp] 
  \centering
  \includegraphics[width=\linewidth]{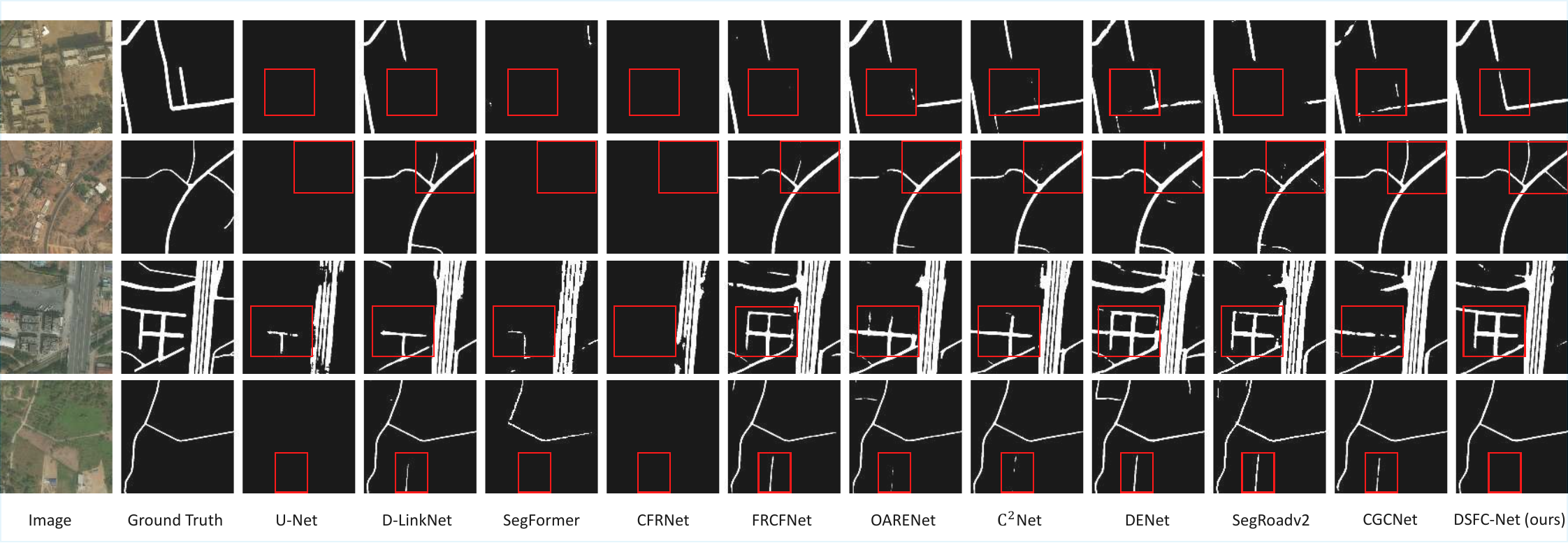}
  \caption{Visual comparison of the different methods on the DeepGlobe dataset.}
  \label{fig5}
\end{figure*}

\subsection{Evaluation Metrics}
The performance of the proposed model is evaluated using precision, recall, F1-score and IoU (Intersection over Union). Precision refers to the percentage of correctly predicted road pixels out of the total predicted pixels, and Recall refers to the percentage of correctly predicted road pixels out of all the actual road pixels. F1-score is a combined metric that balances precision and recall. IoU measures the overlap between the predicted road region and the ground truth road region. The definitions are as follows:

\begin{equation}
    Precision = \frac{TP}{TP + FP},
\end{equation}

\begin{equation}
    Recall = \frac{TP}{TP + FN},
\end{equation}

\begin{equation}
    F1 = 2 \times \frac{Precision \times Recall}{Precision + Recall},
\end{equation}

\begin{equation}
    IoU = \frac{TP}{TP + FP + FN},
\end{equation}

\noindent where $TP$, $FP$, and $FN$ represent the true positive, false positive, and false negative, respectively.

\subsection{Experimental Results}
To comprehensively evaluate the performance of the proposed DSFC-Net on the rural road segmentation task, we conducted a series of experiments on three datasets and compared it with several representative methods in recent years, including U-Net \cite{unet}, D-LinkNet \cite{dlinknet}, SegFormer \cite{segformer}, and newly proposed methods from 2024–2025 such as CFRNet \cite{cfrnet}, FRCFNet \cite{frcfnet}, OARENet \cite{oarenet}, $\text{C}^2$CNet \cite{ccnet}, DENet \cite{denet}, SegRoadv2 \cite{segroadv2}, and CGCNet \cite{cgcnet}. All experiments were conducted under the same data partitioning and training strategy to ensure fair comparison.

\begin{table*}[htbp]
\normalsize
\centering
\caption{Quantitative comparison of the proposed DSFC-Net with the state-of-the-art road extraction methods on the Massachusetts dataset, in which bold is the best and underline is the second-best.}
\label{tab:main3}
\begin{tabular*}{0.7\textwidth}{@{\extracolsep{\fill}}cccccc@{}}
\toprule
\textbf{Methods} &  \textbf{Source}  & \textbf{Precision (\%)} & \textbf{Recall (\%)} & \textbf{F1 (\%)} & \textbf{IoU (\%)} \\ 
\midrule
U-Net \cite{unet} & $\text{MICCAI}_{2015}$  &  85.66  &   72.39  &  78.47  &  64.56  \\
D-LinkNet \cite{dlinknet} & $\text{CVPR}_{2018}$  &  82.81  &   74.38   &  78.37  &  64.43  \\
SegFormer \cite{segformer} & $\text{NIPS}_{2021}$  &  83.51  &  75.02  &  79.04  &  65.34  \\
CFRNet \cite{cfrnet} & $\text{GRSL}_{2024}$  &   85.16  &   75.72  &  80.16  &   66.89  \\
FRCFNet \cite{frcfnet} & $\text{GRSL}_{2024}$  &  85.23   &   75.30   &  79.96  &   66.61  \\
OARENet \cite{oarenet} & $\text{TGRS}_{2024}$  &  84.92   &    75.08   &  79.70  &  66.25  \\
$\text{$\text{C}^2$Net}$ \cite{ccnet} & $\text{TGRS}_{2024}$  &   85.55  &   74.22   &  79.48  &   65.95  \\
DENet \cite{denet} & $\text{TITS}_{2025}$  &  80.30  &   80.56  &  \underline{80.43}  &  \underline{67.26}  \\
SegRoadv2 \cite{segroadv2} & $\text{IJDE}_{2025}$  &  85.41   &  74.27  &  79.45  &  65.91  \\ 
CGCNet \cite{cgcnet} & $\text{TGRS}_{2025}$  & 85.35 & 74.64 &  79.63  &  66.16  \\
\midrule
\textbf{DSFC-Net (Ours)} &  &  85.18  &  76.49  &  \textbf{80.60}  &  \textbf{67.50}  \\ 
\bottomrule
\end{tabular*}
\end{table*}

\begin{figure*}[htbp] 
  \centering
  \includegraphics[width=\linewidth]{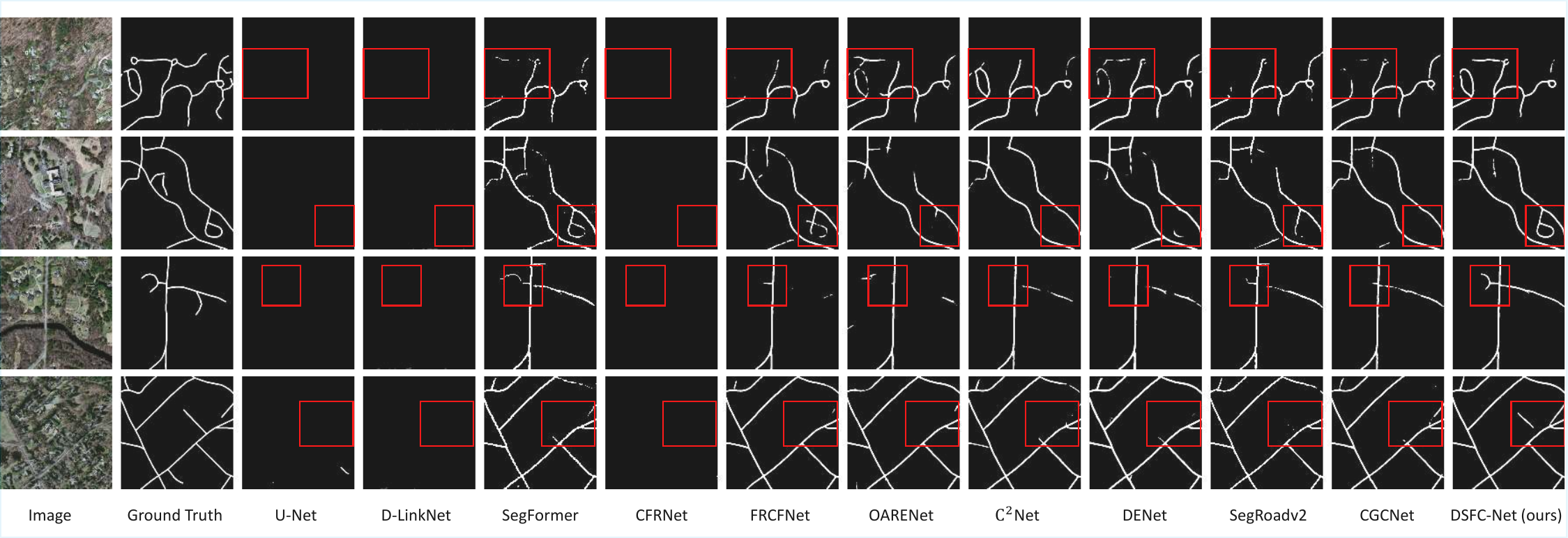}
  \caption{Visual comparison of the different methods on the Massachusetts dataset.}
  \label{fig6}
\end{figure*}

\subsubsection{Results of WHU-RuR+}

The WHU-RuR+ dataset contains a large number of complex rural road scenes with strong background interference and blurred road features, making it extremely challenging. The quantitative results are presented in Table \ref{tab:main1}. DSFC-Net achieves state-of-the-art performance with an F1-score of 69.93\% and an IoU of 53.77\%, outperforming all comparative methods. 

In metric of IoU, our method outperforms the suboptimal D-LinkNet by 1.09\%. In metric of Precision and Recall, recent methods such as $\text{C}^2$Net \cite{ccnet} and CFRNet achieve impressive Precision. However, this comes at the cost of low Recall, indicating that these models fail to detect false negatives objects, such as subtle or obstructed rural roads. Conversely, D-LinkNet achieves the highest Recall but suffers from low Precision, suggesting it is prone to misclassifying background textures as roads. 

DSFC-Net effectively navigates the trade-off between Precision and Recall. We significantly outperform the high-recall D-LinkNet in Precision by 3.31\%, demonstrating our model's ability to suppress noise. Meanwhile, we surpass the high-precision $\text{C}^2$Net in Recall by a substantial margin of 6.48\%, proving our capability to recover disconnected or faint road features. This robustness confirms that the proposed frequency-aware attention and multi-scale fusion mechanisms are particularly effective in extracting complete road topologies in complex rural environments. 

To further validate the effectiveness of DSFC-Net, we present a visual comparison on the WHU-RuR+ dataset in Fig. \ref{fig4}. The visual results demonstrate that our method consistently produces more complete and accurate road segmentation masks compared to state-of-the-art baselines. As illustrated in the second row, which features complex intersections and branching patterns, DSFC-Net preserves topological integrity and maintains smooth connectivity between road segments. In contrast, models like SegFormer and CFRNet exhibit over-smoothing or false positives in junction regions, likely due to their reliance on global attention without explicit frequency-aware boundary refinement.

\subsubsection{Results of DeepGlobe}

The quantitative results of the DeepGlobe dataset are shown in Table \ref{tab:main2}. Overall, the proposed DSFC-Net demonstrates the best comprehensive performance. Specifically, DSFC-Net achieves the F1 score of 81.44\% and the IoU of 68.68\%, both significantly outperforming other comparative methods. Notably, our method outperforms the second-best method CFRNet by a margin of 1.12\% in F1-score and 1.56\% in IoU. Even as the proportion of rural scenes in the DeepGlobe dataset decreases, our proposed model still maintains robust performance across key metrics, demonstrating its strong generalization ability across both urban and rural scenarios.

Fig. \ref{fig5} shows the visualization results of road extraction by the different methods on the DeepGlobe dataset. It is worth noting that in the example on the fourth line, almost all methods incorrectly detected the road below the image, but DSFC Net still maintains robustness in complex scenes.

\subsubsection{Results of Massachusetts}

Serving as a dataset for mixed-scene complexity, the Massachusetts dataset is characterized by a significant proportion of urban road networks (approx. 50\% non-rural background). While these roads are not as narrow and sparse as those in purely rural road datasets, they present distinct challenges, particularly severe occlusion by trees and buildings and interference from similar man-made textures. Therefore, evaluating our model on this dataset is crucial for validating its generalization capability beyond rural-specific domains.

As shown in Table \ref{tab:main3}, DSFC-Net achieves the highest overall performance among all evaluated methods, with an F1-score of 80.60\% and an IoU of 67.50\%. In comparison, DENet employs an aggressive prediction strategy, using edge images as auxiliary features for segmentation. While this strategy improves recall, it also introduces a significant amount of false detection noise due to the complex urban background. Crucially, although the numerical lead in F1-score is marginal, DSFC-Net demonstrates superior reliability with a high Precision of 85.18\%, effectively distinguishing roads from urban noise and ensuring a low false alarm rate.

Fig. \ref{fig6} shows the visualization results demonstrating this robustness. In the first row, the scene contains winding roads embedded in dense vegetation and shadowed regions. Most baseline models fail to detect thin road segments due to limited contextual awareness or are confused by building shadows. DSFC-Net successfully preserves both continuity and topology by leveraging long-range dependencies through the Spatial-Frequency Hybrid Transformer (SFT). In contrast, transformer-based approaches like SegFormer tend to produce fragmented or over-smoothed outputs, particularly at sharp bends and intersections, indicating insufficient boundary refinement in these complex environments.




\begin{table*}[!htbp]
\centering
\normalsize
\caption{Comparison of average performance on three datasets, model complexity and computational cost using average F1-score (Avg F1), average IoU (Avg IoU), parameters (Params) and FLOPs.}
\label{tab:param}
\begin{tabular*}{0.7\textwidth}{@{\extracolsep{\fill}}ccccc@{}}
\toprule
\textbf{Methods} & \textbf{Avg F1 (\%)} & \textbf{Avg IoU (\%)} & \textbf{FLOPs (G)} & \textbf{Params (M)} \\ 
\midrule
U-Net \cite{unet} & 74.57 & 59.78 & 456.87 & 28.99 \\
D-LinkNet \cite{dlinknet} & 75.64 & 61.03 & 80.35 & 44.81 \\
SegFormer \cite{segformer} & 75.27 & 60.66 & 127.61 & 27.34 \\
CFRNet \cite{cfrnet} & 75.60 & 61.21 & 85.23 & 33.64 \\
FRCFNet \cite{frcfnet} & 72.87 & 57.92 & 85.14 & \textbf{7.89} \\
OARENet \cite{oarenet} & 75.54 & 61.07 & 78.85 & 33.48 \\
$\text{$\text{C}^2$Net}$ \cite{ccnet} & 74.98 & 60.25 & 278.44 & 59.61 \\
DENet \cite{denet} & 74.23 & 59.58 & 357.92 & 25.41 \\
SegRoadv2 \cite{segroadv2} & 75.58 & 61.08  & 282.01 & 33.55 \\
CGCNet \cite{cgcnet} & 75.81 & 61.36  & \textbf{62.63} & 8.45 \\
\midrule
\textbf{DSFC-Net (Ours)} & \textbf{77.32} & \textbf{63.32} & 122.57 & 58.06 \\
\bottomrule
\end{tabular*}
\end{table*}

\subsection{Ablation Study and Discussion}

To investigate the effectiveness of the proposed DSFC-Net, we conducted a series of ablation studies on the challenging WHU-RuR+ dataset. We analyze the contribution of the dual-branch architecture, the internal components of the SFT module, and the impact of different CNN backbones.

\subsubsection{Comparison of Model Complexity and Computational Cost}


To assess the trade-off between performance and efficiency, we compare model parameters (Params) and FLOPs. Crucially, FLOPs are computed using the identical $768 \times 768$ input resolution employed in our accuracy benchmarks to ensure rigorous consistency. As shown in Table \ref{tab:param}, DSFC-Net achieves superior accuracy (Avg F1: 77.32\%, Avg IoU: 63.32\%) while maintaining competitive computational efficiency against state-of-the-art methods.

Relative to computationally intensive methods such as U-Net, DENet, C$^2$Net, and SegRoadv2, DSFC-Net shows superior efficiency. For instance, it outperforms the second-best SegRoadv2 by 1.74\% in Avg F1 while reducing FLOPs by 56.5\%. The efficiency of DSFC-Net is primarily attributed to its dual-encoder design and adaptive fusion mechanism, which work in concert to reduce redundant computations.

In contrast, while lightweight architectures like FRCFNet and CGCNet excel in reducing Params and FLOPs, our empirical results demonstrate that a reasonable increase in computational resources translates into significant performance improvements. For instance, compared to the efficient CGCNet, DSFC-Net leverages its additional capacity to achieve a 1.51\% gain in Avg F1 and 1.96\% in Avg IoU. This indicates that the extra parameters and computational cost are effectively utilized to resolve complex segmentation challenges, such as narrow and occluded rural roads.

Finally, compared to the Transformer-based baseline SegFormer, DSFC-Net delivers higher accuracy with slightly lower computational cost. In general, DSFC-Net provides an optimal balance between accuracy and efficiency, suitable for large-scale rural road extraction.

\subsubsection{Impact of Dual-Branch Architecture and Fusion}

We first validate the necessity of the dual-branch design and the Channel Feature Fusion Module (CFFM). As shown in Table \ref{tab:ablation1}, we construct three variants: removing the Transformer branch (w/o Transformer), removing the CNN branch (w/o CNN), and replacing CFFM with simple sum concatenation (w/o CFFM). 

Removing the Transformer branch leads to the most significant performance drop, with F1 decreasing by 1.83\% and IoU decreasing by 2.14\%. This confirms that the long-range semantic dependencies modeled by the Transformer are crucial for identifying continuous road structures in complex environments. Removing the CNN branch results in a slight decrease in F1 and IoU, indicating that while the Transformer is powerful, the local texture features extracted by CNN are still essential for refining boundaries. The replacement of CFFM yields suboptimal results compared to the full model, verifying that our cross-branch joint recalibration strategy is superior to naive feature aggregation.

\begin{table}[!htbp]
\centering
\normalsize
\caption{Ablation Study for the Dual-Branch Architecture and Fusion mechanism on WHU-RuR+ dataset.}
\label{tab:ablation1}
\begin{tabular}{p{0.55\linewidth}cc}
\toprule
\textbf{Methods} & \textbf{F1 (\%)} & \textbf{IoU (\%)} \\ 
\midrule
w/o Transformer & 68.10 & 51.63 \\
w/o CNN & 69.56 & 53.33 \\
w/o CFFM & 69.63 & 53.41 \\
\textbf{DSFC-Net (Ours)} & \textbf{69.93} & \textbf{53.77} \\
\bottomrule
\end{tabular}
\end{table}

\subsubsection{Analysis of SFT Components}

To evaluate the internal mechanisms of the Spatial-Frequency Hybrid Transformer (SFT), we ablate its three core components: Spatial Context Aggregator (SCA), Cross-Frequency Interaction Attention (CFIA), and Multi-scale FFN (MFFN). For SCA and CFIA, we remove them directly, and for MFFN, we replace them with a simple FFN. The results are summarized in Table \ref{tab:ablation2}.

When deployed individually, the SCA module outperforms the CFIA module, suggesting that spatial context is the foundational element for road extraction. Combining CFIA and SCA yields a significant boost, demonstrating that frequency and spatial domains are complementary. CFIA effectively captures high-frequency edge details that SCA might miss. The addition of MFFN further improves performance regardless of the attention mechanism used. The full configuration achieves the highest F1 of 69.93\%, confirming that enhancing local feature transformation within the feed-forward stage is beneficial.

\begin{table}[!htbp]
\centering
\normalsize
\caption{Ablation Study for the components of Spatial-Frequency Hybrid Transformer (SFT) on WHU-RuR+ dataset.}
\label{tab:ablation2}
\begin{tabular}{p{0.16\linewidth}<{\centering}p{0.16\linewidth}<{\centering}p{0.16\linewidth}<{\centering}cc}
\toprule
\multicolumn{3}{c}{\textbf{Modules}} & \multirow{2}{*}{\textbf{F1 (\%)}} & \multirow{2}{*}{\textbf{IoU (\%)}} \\
\cmidrule(lr){1-3}
CFIA & SCA & MFFN & & \\
\midrule
\checkmark & \checkmark & - & 69.23 & 52.94 \\
- & \checkmark & \checkmark & 68.95 & 52.61 \\
\checkmark & - & \checkmark & 68.26 & 51.81 \\
\checkmark & - & - & 68.05 & 51.57 \\
- & \checkmark & - & 68.76 & 52.39 \\
\checkmark & \checkmark & \checkmark & \textbf{69.93} & \textbf{53.77} \\
\bottomrule
\end{tabular}
\end{table}

\subsubsection{Selection of CNN Backbones}

We investigate the impact of different CNN backbones on the overall performance, as shown in Table \ref{tab:ablation3}. We select four of the popular CNN backbones: ResNet \cite{resnet}, MobileNet-v2 \cite{mobilenetv2}, ConvNeXt \cite{convnext} and ConvNeXt-v2 \cite{convnextv2}.

Experiments revealed that MobileNetv2 performed the worst due to its limited feature representation capabilities. ResNet provided a stable baseline performance. ConvNeXt-v2 has the advanced global response normalization design and stronger feature extraction capabilities, surpassing the original ConvNeXt and achieving the best segmentation accuracy. Therefore, we chose ConvNeXt-v2 as the backbone network for the CNN branch.

\begin{table}[!htbp]
\centering
\normalsize
\caption{Ablation Study for the selection of CNN backbones on WHU-RuR+ dataset.}
\label{tab:ablation3}
\begin{tabular}{p{0.55\linewidth}cc}
\toprule
\textbf{CNN Backbones} & \textbf{F1 (\%)} & \textbf{IoU (\%)} \\
\midrule
ResNet & 69.08 & 52.76 \\
MobileNet-v2 & 68.46 & 52.04 \\
ConvNeXt & 69.65 & 53.43 \\
\textbf{ConvNeXt-v2 (Ours)} & \textbf{69.93} & \textbf{53.77} \\
\bottomrule
\end{tabular}
\end{table}

\begin{figure*}[htbp] 
  \centering
  \includegraphics[width=0.9\linewidth]{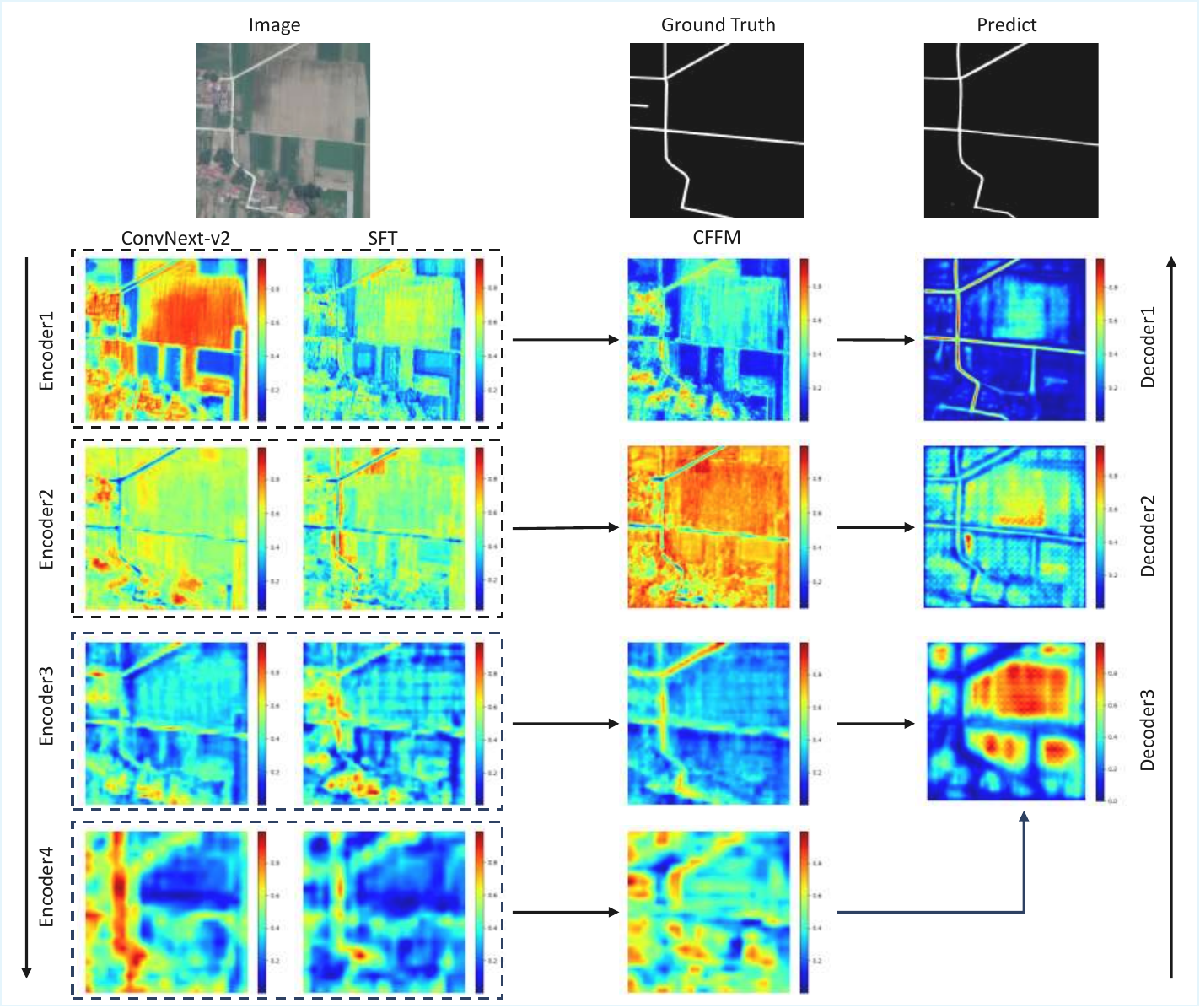}
  \caption{Heatmaps of the DSFC-Net, including the output of the dual encoder, CFFM and decoder.}
  \label{fig:heatmap}
\end{figure*}

\subsubsection{Heatmap Analysis}

To verify the effectiveness of the model more comprehensively, we visualize the heatmaps of the model to analyze the impact of key components. The heatmaps display the output of each layer of encoders, decoders and various CFFM. We use the bilinear interpolation to convert the heatmaps to the same size. As shown in Fig. \ref{fig:heatmap}, these heatmaps reveal how different components progressively refine road-related features across multiple stages. At the early stages, both ConvNext-v2 and SFT exhibit strong responses to large-scale structures such as fields and buildings, but with notable differences: while ConvNext-v2 focuses on local texture contrasts, the SFT branch already begins to emphasize narrow linear patterns through frequency attention. This indicates that the CFIA module in SFT enhances sensitivity to fine-grained structures even at shallow levels. In deeper layers, the CNN branch gradually captures more contextual information about road connectivity and junctions, whereas the SFT module continues to suppress background noise and sharpen road boundaries by attending to dominant spectral frequencies associated with linear objects. The output of CFFM highlights the complementarity of these two branches. After adaptive fusion, the feature maps show enhanced consistency between high-resolution details and low-resolution context, with clear activation along road segments and reduced false activations in non-road regions. In the decoder, the fused features are progressively refined, leading to a final prediction that accurately preserves both the continuity and topology of rural roads. Notably, the decoder’s heatmap shows stronger activation along road edges compared to earlier stages, demonstrating effective propagation of discriminative signals from the dual-branch encoder through the fusion mechanism.

These visualizations confirm that the integration of frequency-domain reasoning via SFT significantly improves the model's ability to distinguish narrow roads from complex backgrounds. And CNN branches can effectively capture local texture details of rural roads. Furthermore, the CFFM successfully balances multi-scale information, ensuring that both local texture cues and global structural context are preserved throughout the model. These key components work together to enable the model to better restore the road structure.

\section{Conclusion}
Rural road extraction from remote sensing imagery mainly confronts three unique challenges, which are 1) high intra-class variability and low inter-class separability, 2) road discontinuity caused by obstruction, and 3) narrow road shapes. 
To tackle these unique challenges, we propose DSFC-Net, a novel dual-encoder architecture that synergistically fuses spatial and frequency-domain information.
A key innovation is the Spatial-Frequency Hybrid Transformer (SFT) layer, tailored to model the global contextual dependencies of winding rural roads. Equipped with a Spatial Context Aggregator (SCA) and Cross-Frequency Interaction Attention (CFIA), the SFT aggregates local spatial context via the SCA while the CFIA analyzes  cross-frequency features through Laplacian Pyramid decomposition, enabling accurate perception of narrow roads and preserving their spatial continuity.
Furthermore, our proposed Channel Feature Fusion Module (CFFM) provides an adaptive mechanism to fuse features from the CNN and Transformer branches, ensuring a robust and comprehensive representation. 
Comprehensive experiments on three public and challenging datasets (WHU-RuR+, DeepGlobe and Massachusetts) validate the superiority of DSFC-Net. It achieves state-of-the-art performance and exhibits strong cross-dataset generalization, effectively extracting complete and accurate rural road networks.
This work offers an effective new tool for rural infrastructure monitoring, which is vital to advancing key Sustainable Development Goals.
While DSFC-Net demonstrates promising performance, there are areas for further optimization.
The current dual-encoder design, while effective for capturing multi-domain features, introduces higher computational demands compared to lightweight CNN-based models. This may pose practical considerations for deployment in resource-constrained edge devices or large-scale operational systems. Future research will explore dynamic frequency selection or sparse attention mechanisms to balance computational efficiency and segmentation accuracy.

\bibliographystyle{IEEEtran}
\bibliography{references}

\newpage

 




\vfill

\end{document}